%% file: main.tex
\def\year{2021}\relax
\documentclass[letterpaper]{article} 
\usepackage{aaai21}  
\usepackage{times}  
\usepackage{helvet} 
\usepackage{courier}  
\usepackage[hyphens]{url}  
\usepackage{bm}
\usepackage{graphicx} 
\def\UrlFont{\rm}  
\usepackage{natbib}  
\usepackage{caption} 
\usepackage{mathtools}  
\usepackage{amsfonts}  
\usepackage{xspace}
\usepackage{float}
\usepackage{color}

\usepackage{threeparttable}
\usepackage{booktabs}

\title{AT-BERT: Adversarial Training BERT for Acronym Identification\\
Winning Solution for SDU@AAAI-21}
\author{
    Danqing Zhu, Wangli Lin, Yang Zhang, Qiwei Zhong, Guanxiong Zeng, Weilin Wu, Jiayu Tang\\
}
\affiliations{
    Alibaba Group, Hangzhou, China\\

    \{danqing.zdq, wangli.lwl, zy142206, yunwei.zqw,
    moshi.zgx, william.wwl, jiayu.tangjy\}@alibaba-inc.com

}
\begin{document}

\maketitle

\begin{abstract}
Acronym identification focuses on finding the acronyms and the phrases that have been abbreviated, which is crucial for scientific document understanding tasks. However, the limited size of manually annotated datasets hinders further improvement for the problem. Recent breakthroughs of language models pre-trained on large corpora clearly show that unsupervised pre-training can vastly improve the performance of downstream tasks. In this paper, we present an \textbf{A}dversarial \textbf{T}raining \textbf{BERT} method named \textbf{AT-BERT}, our winning solution to acronym identification task for Scientific Document Understanding (SDU) Challenge of AAAI 2021. Specifically, the pre-trained BERT is adopted to capture better semantic representation. Then we incorporate the FGM adversarial training strategy into the fine-tuning of BERT, which makes the model more robust and generalized. Furthermore, an ensemble mechanism is devised to involve the representations learned from multiple BERT variants. Assembling all these components together, the experimental results on the SciAI dataset show that our proposed approach outperforms all other competitive state-of-the-art methods.
\end{abstract}

\input{1_introduction}

\input{2_relatedwork}

\input{3_method}

\input{4_experiments}

\input{5_conclusion}

\input{6_acknowledge}

\bibliography{references}

\end{document}

%% file: 1_introduction.tex
\section{Introduction}
\label{sec:introduction}
Acronyms are widespread used in many technical documents to reduce duplicate references to the same concept. 
According to the reports~\cite{barnett2020meta}, after an analysis of more than 24 million article titles and 18 million article abstracts published between 1950 and 2019, there was at least one acronym in 19\% of the titles and 73\% of the abstracts.
As the growing amount of scientific papers published every year, the number of acronyms is also constantly climbing. However, not all acronyms are standard written (i.e., take the first letter of each word and put them together in all capital letters), there are many different ways of writing, e.g., XGBoost is an acronym of eXtreme Gradient Boosting \cite{chen2016xgboost}.
Thus, automatic identification of acronyms and discovery of associated definitions are crucial for text understanding tasks, such as question answering \cite{ackermann2020resolution,veyseh2016cross}, slot filling \cite{pouran2019improving} and definition extraction \cite{kang2020document}.   

\begin{table*}[h]
\renewcommand\arraystretch{1.25}
\centering
\begin{tabular}{ll}  
\hline
\textbf{Input:}   &   Existing methods for learning with noisy labels (LNL) primarily take a loss correction approach. \\
\hline
\textbf{Output:}  &   Existing methods for \underline{learning with noisy labels} (\textbf{LNL}) primarily take a loss correction approach. \\
\hline
\end{tabular}
\caption{A toy example of the acronym identification task. In this example, the acronym is shown in bold font and the long-form is shown with an underline.}
\label{tab:toyexample}
\end{table*}


Several approaches have been proposed to solve the acronym identification problem in the last two decades. The majority of the prior methods are rule-based \cite{schwartz2002simple,okazaki2006building} or feature-based \cite{kuo2009bioadi,liu2017multi}, which employs manually designed rules or features for the acronym and long form predictions. Due to the rules/features are specially designed for finding long forms, these methods have high precision. However, they fail to capture all the diverse forms of acronym expression \cite{harris2019my}. On the contrast, taking advantage of pre-trained word embeddings and deep architecture, deep learning models like LSTM-CRF show promising results for acronym identification \cite{dataset}. Although these works have made great progress, there are still some limitations that hinder further improvement, such as the limited size of manually annotated acronyms and the noises in the automatically created datasets.

Motivated by the above observations, the first publicly available and the largest manually annotated acronym identification the dataset in scientific domain is released \cite{dataset}, and the Scientific Document Understanding (SDU) Challenge~\cite{veyseh2020acronym} for acronym identification task is hosted \footnote{\url{https://sites.google.com/view/sdu-aaai21/shared-task}}. The task aims to identify acronyms (i.e., short-forms) and their meanings (i.e.,long-forms) from the documents, a toy example is shown in Table \ref{tab:toyexample}. In this paper, we formulate the problem as a sentence-level sequence labeling problem, and design a novel BERT-based ensemble model called \textbf{A}dversarial \textbf{T}raining \textbf{BERT} (\textbf{AT-BERT}). Specifically, considering the training data is relatively small, we adopt the pre-trained BERT model as sentence encoder, which is pre-trained on general domain corpora and shows a significant improvement on the performance of downstream tasks with supervised fine-tuning \cite{Beltagy2019SciBERT}. Furthermore, we leverage the FGM \cite{fgm}, an adversarial training strategy to improve the generalization ability of the model, making it more robust to noisy data. Finally, we utilize a multi-BERT ensemble to fully exploit the representations learned from multiple BERT variants \cite{xu2020improving}. Combining these respective advantages, our proposed model won the first prize in the SDU@AAAI-21, outperforming all the other competitive methods.

The main contributions are summarized as follows: 
\begin{itemize}
\item To the best of our knowledge, it is the first work to incorporate adversarial training strategy into BERT-based model for acronym identification task in the scientific domain.
\item We propose a novel framework for acronym identification, including a pre-trained BERT for the semantic representation, an adversarial training strategy to make the model more robust and generalized, as well as a multi-BERT ensemble mechanism to achieve superior performance.
\item Extensive experiments are conducted on the data offered by the SDU@AAAI-21, demonstrating the effectiveness of our proposed method.
\end{itemize}


%% file: 2_relatedwork.tex
\begin{figure*}[!h]
    \centering
    \includegraphics[width=\linewidth]{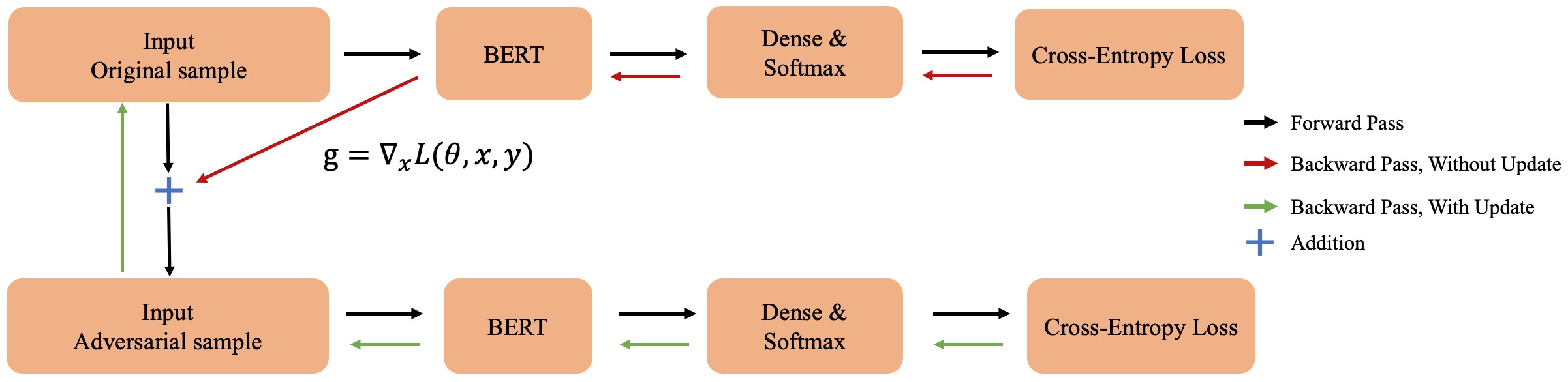}
    \caption{The overall architecture of the proposed AT-BERT approach.}
    \label{fig:figure_net_aaai2021}
\end{figure*}

\section{Related Work}
\label{sec:relatedwork}

In this section, we mainly introduce the related studies for the sequence labeling problem especially the BERT-based models, then we review the existing researches on adversarial training.

\subsection{Sequence Labeling and BERT-based Models}
In this paper, we formulate the acronym identification as a sequence labeling problem.
Traditional approaches of sequence labeling are mainly based on rule-based or feature-based methods \cite{okazaki2006building,kuo2009bioadi}. Recently, deep learning models have achieved promising results, for instance, the LSTM-CRF \cite{ner-survey} model utilizes LSTMs to extract contextualized representations and implement sequence optimization by CRF. With the development of pre-trained language models, BERT-based models achieve state-of-the-art results in natural language related tasks. 
BERT \cite{bert} is a multi-layer bidirectional Transformer encoder, which is pre-trained on Wikipedia and BooksCorpus, has given state-of-the-art results on a wide variety of NLP tasks and inspired many variants. RoBERTa \cite{roberta} utilizes BPE(Byte Pair Encoding) and dynamic masking to increase the shared vocabulary. It optimizes the training strategy of BERT and achieves better performance. ALBERT \cite{albert} utilizes factorized embedding parameterization and cross-Layer parameter sharing to reduce the model parameters. ERINE \cite{ernie} proposes a new masking strategy based on phrases and entities, in which customized tasks are continuously introduced and trained through multi-task learning. 

As for acronym identification, it is more challenging than general sequence labeling problems since acronyms are diverse and ambiguous. Thus the con-textualized representations are crucial and BERT-based models with better semantic representation are more suitable for the task.

\subsection{Adversarial Training}
Adversarial training, in which a network is trained on adversarial examples, is an important way to enhance the robustness of neural networks. 
The Fast Gradient Sign Method (FGSM) \cite{fgsm} and its variant Fast Gradient Method (FGM) \cite{fgm} are firstly proposed for adversarial training. 
The FGSM and FGM methods generate adversarial examples by adding gradient-based perturbation to the input samples with different normalization strategies. They relies heavily on the assumption that the loss function is linear. Different from them, the Projected Gradient Descent (PGD) \cite{pgd} method is an iterative attack method with multi-step iterations, and each iteration will project the perturbation to a specified range. PGD increases computational cost to get better effect, and many PGD-based methods have been proposed to be more efficient. YOPO \cite{yopo} computes the gradient of first layer merely, while FreeAT \cite{free-at} and FreeLB \cite{free} further reduce the frequency of the gradient computation. 

Considering the dataset for acronym identification is relatively small that is easily to be overfit, we incorporate the adversarial training strategy into the BERT-based models to achieve a more robust and generalized performance.

%% file: 3_method.tex
\section{Methodology}
\label{sec:methodology}
In this section, we present the overall architecture of our proposed method, which uses the BERT-based model to solve the sequence labeling problem, and adopt adversarial training strategies to improve the robustness of the model.

\subsection{Overview}
In the following, we propose a BERT-based classification model based on adversarial training strategy, which is called adversarial training BERT (AT-BERT). As show in Figure \ref{fig:figure_net_aaai2021}, the pre-trained BERT model is used for semantic feature encoding, and downstream acronym identification task is solved using its output representations with linear classifiers. In addition, due to the complexity of the acronyms in scientific documents and the relatively small training dataset, the model is prone to overfitting. We use the FGM to add perturbation to the input samples for adversarial training, making the model more robust and generalized. Finally, in order to improve the accuracy of the task, We train different BERT models, such as BERT, SciBERT, RoBERTa, ALBERT and ELECTRA, and make an average ensemble for all the models to achieve superior performance.

\subsection{BERT For Sequence Labeling Problem}
BERT (Bidirectional Encoder Representations from Transformers) is state of the art language model for NLP. It uses the encoder structure of the Transformer \cite{vaswani2017attention} for deep self-supervised learning, which requires task-specific fine-tuning. Transformer is an attention mechanism that learns contextual relations between words (or sub-words) in a text. In this paper, the downstream task is a single sentence tagging problem. We denote a sequence with $T$ words as : $W=(w^1,w^2,...,w^T)$, and a corresponding target as $Y=(y^1,y^2,...,y^T)$. BERT trains an encoder that generates a contextualized vector representation for each token as a hidden state: 
\begin{equation}
\begin{aligned}
H&=\bm{\mathrm{BERT}}(w^1,w^2,...,w^T;\theta) \\
&=(h^1,h^2,...,h^T)
\end{aligned}
\end{equation}
The hidden state is then fed into a fully connection layer with a softmax unit to obtain the predicted probability distribution for each token. The model is trained with the Cross-Entropy loss function, which is defined as follows.
\begin{equation}
L = -\sum_{i}^{T}\sum_{j}^{C} y_j^i log s_j^i  \label{loss}
\end{equation}
where $y^i$ and $s^i$ are the ground truth probability distribution and the predicted probability distribution, $C$ is the number of categories.

\subsection{Adversarial Training For BERT}
Adversarial training is an important way to enhance the robustness of models by adversarial samples. An adversarial example is an instance with small, intentional feature perturbations that induce the model to make a false prediction. 
In the procedure of adversarial training, the input samples will firstly be mixed with some small perturbation to generate adversarial samples~\cite{szegedy2013intriguing}. 
The model is trained with both the original input sample and generated adversarial samples to enhance its robustness and  generalization.
\cite{pgd} abstracted the general form of adversarial training as the maximum and minimum formula as follows.
\begin{equation}
\min_\theta \mathbb{E}_{(x, y)\sim \mathcal{D}} [\max_{r_{adv} \in \mathcal{S}} L(\theta, x+r_{adv}, y)]  \label{min_max}
\end{equation}
where $x$ represents the input representation of the sample, and a corresponding target as $y$, $r_{adv}$ is the perturbation applied to the input, $\mathcal{S}$ is the perturbation space, and $L$ is some loss function like Equation~(\ref{loss}). First, The internal maximization problem is to find the best perturbation at a given data point $x$ in the perturbation space to generate adversarial examples that achieves high loss. This can be seen as an attack on a given neural network. Second, The goal of the external minimization problem is to find the model parameters $\theta$ to minimize the ``adversarial loss" given by the internal attack problem.

With the above clear definition of the adversarial training, we will introduce how to apply a small perturbation to the input sample to generate adversarial samples in our task. There are many related studies on adversarial training such as the FGSM, single-step algorithm FGM, multi-step algorithm PGD, and Free-LB. Since these can basically be regarded as a series of methods, we will briefly introduce FGM. FGM made a simple extension on the calculation of perturbation in FGSM and proposed FGM. The main idea is to add a perturbation to the input that can increase the loss, it happens to be the direction in which the gradient of the loss function rises. Specifically, the adversarial perturbation is  defined as follows.

\begin{equation}
g = \nabla_x L(\theta, x, y)
\end{equation}
\begin{equation}
r_{adv} = \epsilon \cdot \frac{g}{||g||_2} 
\label{adv}
\end{equation}
where $g$ is the gradient of the loss with respect to $x$, the L2 norm is used to constrain $g$ in Equation (\ref{adv}), and the $\epsilon$ is a hyperparameter and defaults to 1. In our acronym identification task, the perturbation $r_{adv}$ will be added to the embedding of the input word. 
The overall architecture of the proposed AT-BERT is shown in Figure \ref{fig:figure_net_aaai2021}.

%% file: 4_experiments.tex
\section{Experiments}
\label{sec:experiments}

In this section, we first introduce the experimental dataset and evaluation metrics, and then conduct comprehensive experimental studies to verify the effectiveness of our method.

\begin{table}
\centering
\begin{tabular}{c|cc}
\hline
\textbf{Data} & \textbf{Sample Number} & \textbf{Ratio} \\ 
\hline
\textbf{training set}       & 14,006        & 80.16\%         \\ 
\textbf{development set}      & 1,717        & 9.82\%         \\ 
\textbf{test set}       & 1,750       & 10.02\%         \\ \hline
\textbf{total}      & 17,473         & 100\%         \\ \hline
\end{tabular}
\caption{The statistical information of dataset.}
\label{table:dataset_num}
\end{table} 

\begin{figure}
    \centering
    \includegraphics[scale=0.4]{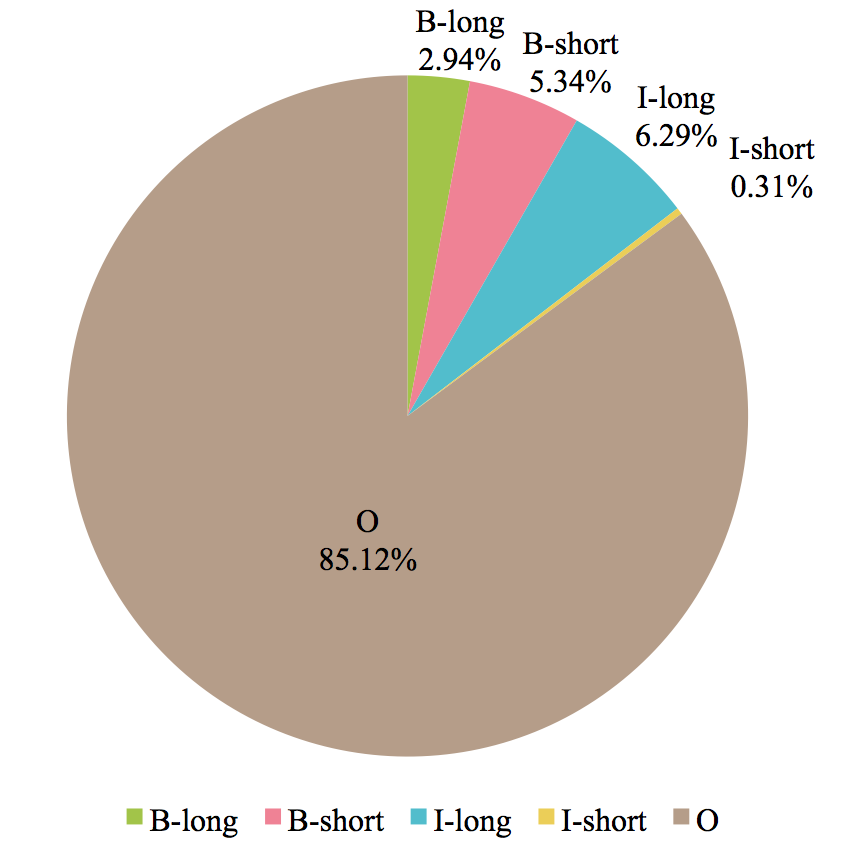}
    \caption{Category distribution of training set.}
    \label{fig:category distribution of train-set}
\end{figure}

\subsection{Dataset}
We evaluate all models based on the dataset provided by SDU@AAAI-21.
It contains a training set of 14,006 samples, a development set of 1,717 samples, and a test set of 1,750 samples, as shown in Table~\ref{table:dataset_num}.
This task aims to identify acronyms (i.e., short-forms) and their meanings (i.e., long-forms) from the documents.
The dataset provides the boundaries for the acronyms and long forms in the sentence using BIO format (i.e., label set includes B-short, I-short, B-long, I-long and O). 
The percentage of each label category in all tokens is shown in Figure~\ref{fig:category distribution of train-set}.
Obviously, the distribution of label classes across the all known classes is biased.
Each sample in the training set and development set has three attributes:
\begin{itemize}
\item \textbf{tokens}: The list of words (tokens) of the sample.
\item \textbf{labels}: The short-form and long-form labels of the words in BIO format. The labels B-short and B-long identifies the beginning of a short-form and long-form phrase, respectively. The labels I-short and I-long indicates the words inside the short-form or long-form phrases. Finally, the label O shows the word is not part of any short-form or long-form phrase.
\item \textbf{id}: The unique ID of the sample.
\end{itemize}
And the test set has no labels attributes.
We refer the readers to the work~\cite{dataset} for more details.

\begin{table*}

\begin{threeparttable}
\resizebox{\linewidth}{!}{
\begin{tabular}{c|l|cccccc}
\hline
{\textbf{Parameter}}                                                                                                                                    & \textbf{SciBERT}                                                            & \textbf{BERT }                                                        & \textbf{BERT}$_\mathrm{LARGE}$                                                  & \textbf{RoBERTa }      & \textbf{ALBERT }           & \textbf{ELECTRA    }                                                                        \\ \hline
{\begin{tabular}[c]{@{}c@{}}
Training\\    Arguments\end{tabular}}      & pretrained\_model                                                          & \begin{tabular}[c]{@{}c@{}}scibert-scivocab\\ -uncased\tnote{a} \end{tabular} & \begin{tabular}[c]{@{}c@{}}bert-base\\ -uncased\tnote{b} \end{tabular} & \begin{tabular}[c]{@{}c@{}}bert-large\\ -uncased\tnote{c} \end{tabular} & roberta-large\tnote{d} & albert-xxlarge-v2\tnote{e} & \begin{tabular}[c]{@{}c@{}}google/electra-large\\      -discriminator\tnote{f} \end{tabular} \\ 
                                                                                      & epoch                                                                      & 3                                                                   & 3                                                            & 3                                                             & 3             & 4                 & 3                                                                                  \\ 
                                                                                      & batch\_size                                                                & 16                                                                  & 16                                                           & 16                                                            & 16            & 8                 & 16                                                                                 \\ 
                                                                                      & learning rate                                                              & 2e-5                                                                & 2e-5                                                         & 2e-5                                                          & 2e-5          & 5e-6              & 2e-5                                                                               \\ 
                                                                                      & max\_seq\_len                                                              & 512                                                                 & 512                                                          & 512                                                           & 512           & 512               & 512                                                                                \\ \hline
{\begin{tabular}[c]{@{}c@{}}Model \\      Arguments\end{tabular}} & \begin{tabular}[c]{@{}l@{}}attention\_probs\\ \_dropout\_prob\end{tabular} & 0.1                                                                 & 0.1                                                          & 0.1                                                           & 0.1           & 0                 & 0.1                                                                                \\ 
                                                                                      & hidden\_dropout\_prob                                                      & 0.1                                                                 & 0.1                                                          & 0.1                                                           & 0.1           & 0                 & 0.1                                                                                \\ 
                                                                                      & classifier\_dropout\_prob                                                  &                                                                     &                                                              &                                                               &               & 0.1               &                                                                                    \\ 
                                                                                      & num\_attention\_heads                                                      & 16                                                                  & 12                                                           & 16                                                            & 16            & 64                & 16                                                                                 \\ 
                                                                                      & num\_hidden\_layers                                                        & 24                                                                  & 12                                                           & 24                                                            & 24            & 12                & 24                                                                                 \\ 
                                                                                      & hidden\_size                                                               & 1024                                                                & 768                                                          & 1024                                                          & 1024          & 4096              & 1024                                                                               \\ 
                                                                                      & hidden\_act                                                                & gelu                                                                & gelu                                                         & gelu                                                          & gelu          & gelu\_new         & gelu                                                                               \\ 
                                                                                      & intermediate\_size                                                         & 3072                                                                & 3072                                                         & 4096                                                          & 4096          & 16384             & 4096                                                                               \\ 
                                                                                      & vocab\_size                                                                & 30522                                                               & 30522                                                        & 30522                                                         & 50265         & 30000             & 30522                                                                              \\ \hline
\end{tabular}
}
\begin{tablenotes}
\item[a] \url{https://github.com/allenai/scibert}
\item[b] \url{https://huggingface.co/bert-base-uncased}
\item[c] \url{https://huggingface.co/bert-large-uncased}
\item[d] \url{https://huggingface.co/roberta-large}
\item[e] \url{https://huggingface.co/albert-xxlarge-v2}
\item[f] \url{https://huggingface.co/google/electra-large-discriminator}
\end{tablenotes}
\end{threeparttable}

\caption{Model architecture and main parameters of our experiments.}
\label{tab:impts}
\end{table*}
\subsection{Evaluation Metrics}
Regarding the evaluation metrics, similar to previous work~\cite{dataset}, the results are evaluated based on their macro-averaged precision, recall, and F1 score on the test set computed for correct predictions of short-form (i.e., acronym) and long-form (i.e., phrase) boundaries in the sentences.
A short-form or long-form boundary prediction is counted as correct if the beginning and the end of the predicted short-form or long-from boundaries equal to the ground-truth beginning and end of the short-form or long-form boundary, respectively.
The official score (noded as MacroF1) is the macro average of short-form and long-form prediction F1 score.

\subsection{Compared Methods}
We experiment with four schemes: Baselines, BERT models, Adversarial Training for BERT (AT-BERT) and Model Ensemble.
\subsubsection{(a) Baselines}
\begin{itemize}
    \item \textbf{Rule-based methods:} These models employ manually designed rules to extract acronyms and long forms in the text. The evaluation code and results are provided by SDU@AAAI-21\footnote{\url{https://github.com/amirveyseh/AAAI-21-SDU-shared-task-1-AI}}.
    \item \textbf{Deep learning models:} As shown in previous work~\cite{dataset}, we can see that the F1 score of the LSTM-CRF model is only one percentage point higher than the rule-based models. Therefore, we do not implement the LSTM-CRF model by ourselves. More details on these models and hyper parameters are illustrated in~\cite{dataset}.
\end{itemize}

\subsubsection{(b) BERT Models } 
\begin{itemize}
    \item \textbf{BERT:} BERT~\cite{bert} is a multi-layer bidirectional transformer encoder trained with a masked language modeling (MLM) objective and the next sentence prediction task. It has two sizes, we have both experimented, namely \(\rm BERT_{BASE}\) architecture (L=12, H=768, A=12, total 110M parameters) and \(\rm BERT_{LARGE}\) architecture (L=24, H=1024, A=16, total 355M parameters) provided by huggingface~\cite{wolf-etal-2020-transformers}.
    \item \textbf{SciBERT:} SciBERT is the pretrained model presented by \citeauthor{Beltagy2019SciBERT}, which is based on \(\rm BERT_{BASE}\) and trained on a large corpus of scientific text. It has achieved new state-of-the-art results on a suite of tasks in the scientific domain
    \cite{Beltagy2019SciBERT,qwzhong2020}.
    \item \textbf{RoBERTa:} RoBERTa~\cite{roberta} improves the original implementation of BERT for better performance, using dynamic masking, removing the next sentence prediction task, training with larger batches, on more data, and for longer. RoBERTa follows the same architecture as BERT.
    \item \textbf{ALBERT:} The ALBERT model~\cite{albert} presents two parameter-reduction techniques to lower memory consumption and increases the training speed of BERT. First, splitting the embedding matrix into two smaller matrices. Second, using repeating layers split among groups.
    \item \textbf{ELECTRA:} ELECTRA\cite{clark2020electra} proposes a more effective pretraining method. Instead of corrupting some positions of inputs with [MASK], ELECTRA replaces some tokens of the inputs with their plausible alternatives sampled from a small generator network. ELECTRA trains a discriminator to predict whether each token in the corrupted input was replaced by the generator or not. The pretrained discriminator can then be used in downstream tasks for fine-tuning.
\end{itemize}

\subsubsection{(c) AT-BERT Models }
\ \newline
In order to solve the problem that the models may be overfitted and have poor generalization due to less training data, we used the FGM algorithm for adversarial training on various BERT models.

\subsubsection{(d) Model Ensemble}
\ \newline
Model ensemble is a commonly used method to improve model accuracy. We perform an average ensemble of the output probability distributions of various BERT models to obtain the final prediction results. In general, model fusion requires that the fused models themselves perform well and are different from each other, so we finally use four models: \(\rm BERT_{LARGE}\), RoBERTa, ALBERT, and ELECTRA for fusion (named BERT-E shortly). AT-BERT equipped with adversarial training strategy is shorted as AT-BERT-E.

\subsection{Implementation}
All models are implemented based on the open source transformers library of huggingface \cite{wolf-etal-2020-transformers}, which provides thousands of pretrained models to perform tasks on texts such as sequence classification and information extraction. It provides APIs to quickly download and use those pretrained models on a given text, fine-tune them on your own datasets. The deep learning framework used in this paper is Pytorch. In addition, We use two V100 GPUs with 12 cores to complete these experiments.

For the above models, we do not modify the original network structure. For more detailed network structures and parameters, please refer to transformers\cite{wolf-etal-2020-transformers}. 
For each BERT variant model, we pick the best learning rate and number of epochs on the development set and report the corresponding test results. 
We found that when epoch is set to 3, the learning rate is 2e-5, the maximum sentence length is 512 and the batch size is set to occupy as much GPU as possible, most models are close to convergence.
Therefore, we set the above training parameters uniformly for all models.
More detailed parameter settings are shown in Table~\ref{tab:impts}.

\begin{table*}[]
\begin{tabular}{c|c|ccc|ccc|c}
\hline
\textbf{Scheme}                   & \textbf{Methodology}      & \multicolumn{3}{c|}{\textbf{Acronym}}             & \multicolumn{3}{c|}{\textbf{Long   Form}}         & \textbf{}            \\ 
                                   &                           & \textbf{P(\%)} & \textbf{R(\%)} & \textbf{F1(\%)} & \textbf{P(\%)} & \textbf{R(\%)} & \textbf{F1(\%)} & \textbf{MacroF1(\%)} \\ \hline
{\textbf{Baseline}} & \textbf{RULE}             & 90.67          & 91.71          & 91.18           & 95.78          & 66.09          & 78.21           & 85.46                \\ 
                                   & \textbf{LSTM}-\textbf{CRF}         & 88.58          & 86.93          & 87.75           & 85.33          & 85.38          & 85.36           & 86.55                \\ \hline
{\textbf{BERT}}     & \textbf{BERT}$_\mathrm{BASE}$        & 92.88          & 92.50          & 92.69           & 87.20          & 89.96          & 88.56           & 90.63                \\ 
                                   & \textbf{SciBERT}          & 92.61          & 90.82          & 91.71           & 90.96          & 92.37          & 91.66           & 91.69                \\ 
                                   & \textbf{BERT}$_\mathrm{LARGE}$       & 94.07          & 94.28          & 94.18           & 90.60          & 91.44          & 91.02           & 92.60                \\ 
                                   & \textbf{RoBERTa}          & 93.10          & 92.63          & 92.86           & 92.77          & 93.92          & 93.35           & 93.11                \\ 
                                   & \textbf{ALBERT}           & 91.82          & 94.22          & 93.01           & 91.69          & 94.36          & 93.00           & 93.00                \\ 
                                   & \textbf{ELECTRA}          & 92.79          & 93.99          & 93.39           & 91.25          & 94.42          & 92.81           & 93.10                \\ \hline
{\textbf{AT-BERT}}  & \textbf{BERT}$_\mathrm{LARGE}$       & 94.34          & 93.17          & 93.75           & 92.04          & 93.24          & 92.64           & 93.20                \\ 
                                   & \textbf{RoBERTa}          & 94.50          & 93.36          & 93.93           & 91.83          & 94.73          & 93.26           & 93.60                \\ 
                                   & \textbf{ALBERT}           & 92.48          & 94.01          & 93.24           & 92.73          & 94.44          & 93.56           & 93.41                \\ 
                                   & \textbf{ELECTRA}          & 94.38          & 92.88          & 93.63           & 92.66          & 93.86          & 93.26           & 93.45                \\ \hline
{\textbf{Ensemble}} & \textbf{BERT}-\textbf{E}    & 94.62          & 92.72          & 93.66           & 92.83          & 93.99          & 93.18           & 93.43                \\ 
                                   & \textbf{AT}-\textbf{BERT}-\textbf{E} & \textbf{94.87}          & \textbf{93.99}          & \textbf{94.43}           & \textbf{92.84}          & \textbf{94.79}          & \textbf{93.80}           & \textbf{94.12}                \\ \hline
\textbf{Human Performance}          & \textbf{Human Performance} & 98.51          & 94.33          & 96.37           & 96.89          & 94.79          & 95.82           & 96.09                \\ \hline
\end{tabular}
\caption{Performance comparison over different released period. }
\label{table:experiments_res}
\end{table*}

\begin{table*}[]
\scalebox{0.8}{
\begin{tabular}{c|c|c|c|c|c|c|c|c|c|c|c|c|c|c|c|c|c}
\hline
\textbf{tokens}     & this & study & were & convolutional & and/or & recurrent & neural & nets   & ( & CNNs    & , & RNNs    & , & or & CRNNs   & ) & \textbf{,} \\ \hline
\textbf{label}      & O    & O     & O    & B-long        & I-long & I-long    & I-long & I-long & O & B-short & O & B-short & O & O  & B-short & O & O          \\ \hline
\textbf{w/o AT} & O    & O     & O    & O             & O      & B-long    & I-long & I-long & O & B-short & O & B-short & O & O  & B-short & O & O          \\ \hline
\textbf{with AT}    & O    & O     & O    & B-long        & I-long & I-long    & I-long & I-long & O & B-short & O & B-short & O & O  & B-short & O & O          \\ \hline
\end{tabular}
}
\caption{Case analysis with and without (w/o) adversarial training.}
\label{table:Case analysis}
\end{table*}
\subsection{Performance Comparison}
The comparison results are shown in Table~ \ref{table:experiments_res}. The main observations are summarized as follows:

(1) Compared with the rule-based method and LSTM-CRF model, all the BERT-based models achieve better results, illustrating the advantage with pre-trained BERT.
Due to the conservative nature of rule-base method, it has higher precision but far lower recall than all other models.
With unsupervised pre-training on large corpus, the BERT-based models outperform LSTM-CRF among all the evaluation metrics.

(2) Among the six different BERT-based models, the SciBERT model has the same architecture and training strategy with \(\rm BERT_{BASE}\). However, the SciBERT, whose corpus is more relevant to our task, outperforms \(\rm BERT_{BASE}\) with 1.03\% increased MarcoF1 score. Meanwhile, the \(\rm BERT_{LARGE}\) have more complex architecture and parameter, thus it performs better than SciBERT. Taking advantage of larger training corpus and more effective training strategies, the performances of other BERT-based models like RoBERTa and ELECTRA get further improvement.

\begin{figure}[!h]
    \centering
    \includegraphics[width=\linewidth]{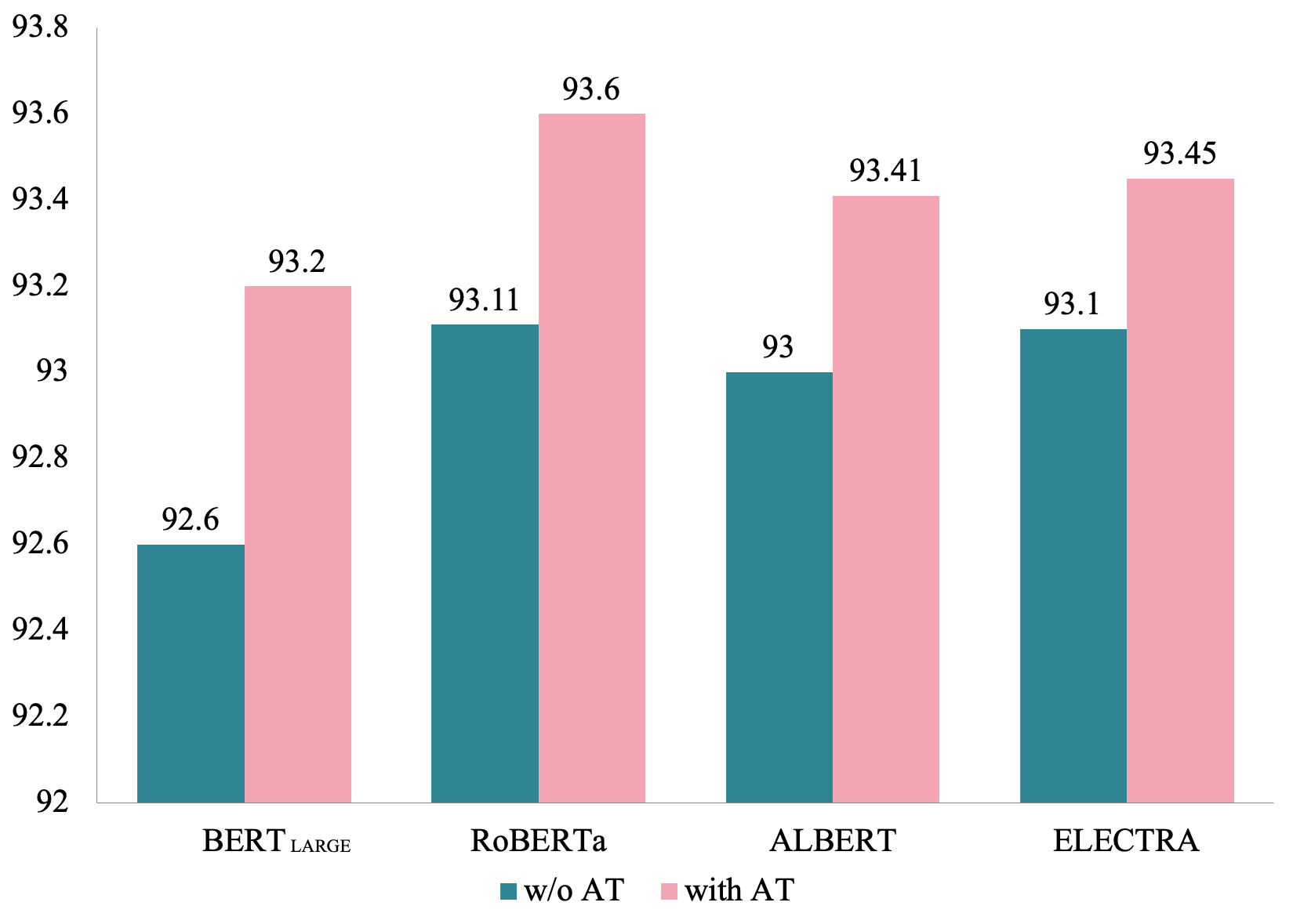}
    \caption{Comparison of Macro F1 for models with and without (w/o) adversarial training.}
    \label{fig:at_model_comp}
\end{figure}

(3) With the FGM adversarial training strategy, as shown in Figure \ref{fig:at_model_comp}, we can clearly observe that the AT-BERT based models outperforms those without adversarial training by a large margin. The obvious improvement indicates that the adversarial training strategy has a positive effect on the BERT-based models' performance.

(4) From the comparison of ensemble strategies, we can find that the BERT-E model is more superior than any BERT-based model, especially in the precision and MarcoF1 metrics. The similar phenomenon also occurs in the comparison of AT-BERT-E model with single AT-BERT based model. The AT-BERT-E model which performs best is more advanced than the baseline methods, i.e., the rule-based method and LSTM-CRF model, with 8.66\% and 7.57\% increased MarcoF1 score, respectively.

The above observations demonstrate that the effectiveness of different components of our proposed AT-BERT model. 
However, the best performance is still less effective than human performance, thereby providing many research opportunities for this scenario.

\subsection{Case Study} 
We further analyze the prediction results of BERT and AT-BERT.
An interesting example (DEV-1629) is shown in Table~\ref{table:Case analysis}. The corresponding long-forms of ``\textit{CNNs}'', ``\textit{RNNs}'', and ``\textit{CRNNs}'' is ``\textit{convolutional} \textit{and/or} \textit{recurrent} \textit{neural} \textit{nets}'', while the prediction of BERT is ``\textit{recurrent} \textit{neural} \textit{nets}''. This example is very confusing, because ``\textit{recurrent} \textit{neural} \textit{nets}'' can be considered as the long form of ``\textit{RNNs}''. The general BERT model might be easily affected by the token ``\textit{and/or}'' and ignores the previous token ``\textit{convolutional}. The experimental results prove that our proposed AT-BERT does have better robustness and generalization.

%% file: 5_conclusion.tex
\section{Conclusion and Future Work}
\label{sec:conclusion}
In this paper, we proposed a novel BERT-based model called AT-BERT for acronym identification, the winning solution to the AAAI-21 Workshop on Scientific Document Understanding. A FGM-based adversarial training strategy was incorporated in the fine-tuning of BERT variants, and an average ensemble mechanism was devised to capture the better representation from multi-BERT variants. The extensive experiments were conducted on the SciAI dataset and achieved the best performance among all the competitive methods, which verifies the effectiveness of the proposed approach.  

In the future, we will optimize our model from two perspectives. One is to explore more adversarial training strategies such as PGD and FreeLB for BERT model. The other is to try different loss function such as Dice Loss \cite{li2019dice} and Focal Loss \cite{lin2017focal} to alleviate the phenomenon of class imbalance.

%% file: 6_acknowledge.tex
\section{Acknowledgments}
\label{sec:acknowledgments}
We thank the organizers of acronym identification and disambiguation competitions and the reviewers for their valuable comments and suggestions.

%% file: main.bbl
\begin{thebibliography}{33}
\providecommand{\natexlab}[1]{#1}
\providecommand{\url}[1]{\texttt{#1}}
\providecommand{\urlprefix}{URL }
\expandafter\ifx\csname urlstyle\endcsname\relax
  \providecommand{\doi}[1]{doi:\discretionary{}{}{}#1}\else
  \providecommand{\doi}{doi:\discretionary{}{}{}\begingroup
  \urlstyle{rm}\Url}\fi

\bibitem[{Ackermann et~al.(2020)Ackermann, Beller, Boxwell, Katz, and
  Summers}]{ackermann2020resolution}
Ackermann, C.~F.; Beller, C.~E.; Boxwell, S.~A.; Katz, E.~G.; and Summers,
  K.~M. 2020.
\newblock Resolution of Acronyms in Question Answering Systems.
\newblock US Patent 10,572,597.

\bibitem[{Barnett and Doubleday(2020)}]{barnett2020meta}
Barnett, A.; and Doubleday, Z. 2020.
\newblock Meta-Research: The Growth of Acronyms in the Scientific Literature.
\newblock \emph{Elife} 9: e60080.

\bibitem[{Beltagy, Lo, and Cohan(2019)}]{Beltagy2019SciBERT}
Beltagy, I.; Lo, K.; and Cohan, A. 2019.
\newblock SciBERT: Pretrained Language Model for Scientific Text.
\newblock In \emph{EMNLP}.

\bibitem[{Chen and Guestrin(2016)}]{chen2016xgboost}
Chen, T.; and Guestrin, C. 2016.
\newblock XGBoost: A Scalable Tree Boosting System.
\newblock In \emph{KDD}, 785–794.

\bibitem[{Clark et~al.(2020)Clark, Luong, Le, and Manning}]{clark2020electra}
Clark, K.; Luong, M.; Le, Q.~V.; and Manning, C.~D. 2020.
\newblock {ELECTRA:} Pre-training Text Encoders as Discriminators Rather Than
  Generators.
\newblock In \emph{ICLR}.

\bibitem[{Goodfellow, Shlens, and Szegedy(2015)}]{fgsm}
Goodfellow, I.~J.; Shlens, J.; and Szegedy, C. 2015.
\newblock Explaining and Harnessing Adversarial Examples.
\newblock In \emph{ICLR}.

\bibitem[{Harris and Srinivasan(2019)}]{harris2019my}
Harris, C.~G.; and Srinivasan, P. 2019.
\newblock My Word! Machine versus Human Computation Methods for Identifying and
  Resolving Acronyms.
\newblock \emph{Computaci{\'o}n y Sistemas} 23(3).

\bibitem[{Kang et~al.(2020)Kang, Head, Sidhu, Lo, Weld, and
  Hearst}]{kang2020document}
Kang, D.; Head, A.; Sidhu, R.; Lo, K.; Weld, D.~S.; and Hearst, M.~A. 2020.
\newblock Document-Level Definition Detection in Scholarly Documents: Existing
  Models, Error Analyses, and Future Directions.
\newblock \emph{arXiv preprint arXiv:2010.05129} .

\bibitem[{Kenton and Toutanova(2019)}]{bert}
Kenton, J. D. M.-W.~C.; and Toutanova, L.~K. 2019.
\newblock BERT: Pre-training of Deep Bidirectional Transformers for Language
  Understanding.
\newblock In \emph{NAACL-HLT}, 4171--4186.

\bibitem[{Kuo et~al.(2009)Kuo, Ling, Lin, and Hsu}]{kuo2009bioadi}
Kuo, C.-J.; Ling, M.~H.; Lin, K.-T.; and Hsu, C.-N. 2009.
\newblock BIOADI: A Machine Learning Approach to Identifying Abbreviations and
  Definitions in Biological Literature.
\newblock In \emph{BMC bioinformatics}, volume~10, S7. Springer.

\bibitem[{Lan et~al.(2019)Lan, Chen, Goodman, Gimpel, Sharma, and
  Soricut}]{albert}
Lan, Z.; Chen, M.; Goodman, S.; Gimpel, K.; Sharma, P.; and Soricut, R. 2019.
\newblock ALBERT: A Lite BERT for Self-supervised Learning of Language
  Representations.
\newblock In \emph{ICLR}.

\bibitem[{Li et~al.(2020)Li, Sun, Han, and Li}]{ner-survey}
Li, J.; Sun, A.; Han, J.; and Li, C. 2020.
\newblock A Survey on Deep Learning for Named Entity Recognition.
\newblock \emph{IEEE Transactions on Knowledge and Data Engineering} .

\bibitem[{Li et~al.(2019)Li, Sun, Meng, Liang, Wu, and Li}]{li2019dice}
Li, X.; Sun, X.; Meng, Y.; Liang, J.; Wu, F.; and Li, J. 2019.
\newblock Dice Loss for Data-imbalanced NLP Tasks.
\newblock \emph{arXiv preprint arXiv:1911.02855} .

\bibitem[{Lin et~al.(2017)Lin, Goyal, Girshick, He, and
  Doll{\'a}r}]{lin2017focal}
Lin, T.-Y.; Goyal, P.; Girshick, R.; He, K.; and Doll{\'a}r, P. 2017.
\newblock Focal Loss for Dense Object Detection.
\newblock In \emph{ICCV}, 2980--2988.

\bibitem[{Liu, Liu, and Huang(2017)}]{liu2017multi}
Liu, J.; Liu, C.; and Huang, Y. 2017.
\newblock Multi-granularity Sequence Labeling Model for Acronym Expansion
  Identification.
\newblock \emph{Information Sciences} 378: 462--474.

\bibitem[{Liu et~al.(2019)Liu, Ott, Goyal, Du, Joshi, Chen, Levy, Lewis,
  Zettlemoyer, and Stoyanov}]{roberta}
Liu, Y.; Ott, M.; Goyal, N.; Du, J.; Joshi, M.; Chen, D.; Levy, O.; Lewis, M.;
  Zettlemoyer, L.; and Stoyanov, V. 2019.
\newblock RoBERTa: A Robustly Optimized BERT Pretraining Approach.
\newblock \emph{arXiv preprint arXiv:1907.11692} .

\bibitem[{Madry et~al.(2018)Madry, Makelov, Schmidt, Tsipras, and Vladu}]{pgd}
Madry, A.; Makelov, A.; Schmidt, L.; Tsipras, D.; and Vladu, A. 2018.
\newblock Towards Deep Learning Models Resistant to Adversarial Attacks.
\newblock In \emph{ICLR}.

\bibitem[{Miyato, Dai, and Goodfellow(2017)}]{fgm}
Miyato, T.; Dai, A.~M.; and Goodfellow, I.~J. 2017.
\newblock Adversarial Training Methods for Semi-Supervised Text Classification.
\newblock In \emph{ICLR}.

\bibitem[{Okazaki and Ananiadou(2006)}]{okazaki2006building}
Okazaki, N.; and Ananiadou, S. 2006.
\newblock Building an Abbreviation Dictionary Using a Term Recognition
  Approach.
\newblock \emph{Bioinformatics} 22(24): 3089--3095.

\bibitem[{Pouran Ben~Veyseh, Dernonrcourt, and
  Nguyen(2019)}]{pouran2019improving}
Pouran Ben~Veyseh, A.; Dernonrcourt, F.; and Nguyen, T.~H. 2019.
\newblock Improving Slot Filling by Utilizing Contextual Information.
\newblock \emph{arXiv} arXiv--1911.

\bibitem[{Schwartz and Hearst(2002)}]{schwartz2002simple}
Schwartz, A.~S.; and Hearst, M.~A. 2002.
\newblock A Simple Algorithm for Identifying Abbreviation Definitions in
  Biomedical Text.
\newblock In \emph{Biocomputing 2003}, 451--462. World Scientific.

\bibitem[{Shafahi et~al.(2019)Shafahi, Najibi, Ghiasi, Xu, Dickerson, Studer,
  Davis, Taylor, and Goldstein}]{free-at}
Shafahi, A.; Najibi, M.; Ghiasi, M.~A.; Xu, Z.; Dickerson, J.; Studer, C.;
  Davis, L.~S.; Taylor, G.; and Goldstein, T. 2019.
\newblock Adversarial Training for Free!
\newblock In \emph{NeurIPS}, 3358--3369.

\bibitem[{Sun et~al.(2019)Sun, Wang, Li, Feng, Chen, Zhang, Tian, Zhu, Tian,
  and Wu}]{ernie}
Sun, Y.; Wang, S.; Li, Y.; Feng, S.; Chen, X.; Zhang, H.; Tian, X.; Zhu, D.;
  Tian, H.; and Wu, H. 2019.
\newblock Ernie: Enhanced Representation Through Knowledge Integration.
\newblock \emph{arXiv preprint arXiv:1904.09223} .

\bibitem[{Szegedy et~al.(2014)Szegedy, Zaremba, Sutskever, Bruna, Erhan,
  Goodfellow, and Fergus}]{szegedy2013intriguing}
Szegedy, C.; Zaremba, W.; Sutskever, I.; Bruna, J.; Erhan, D.; Goodfellow,
  I.~J.; and Fergus, R. 2014.
\newblock Intriguing Properties of Neural Networks.
\newblock In Bengio, Y.; and LeCun, Y., eds., \emph{ICLR}.

\bibitem[{Vaswani et~al.(2017)Vaswani, Shazeer, Parmar, Uszkoreit, Jones,
  Gomez, Kaiser, and Polosukhin}]{vaswani2017attention}
Vaswani, A.; Shazeer, N.; Parmar, N.; Uszkoreit, J.; Jones, L.; Gomez, A.~N.;
  Kaiser, {\L}.; and Polosukhin, I. 2017.
\newblock Attention is All You Need.
\newblock In \emph{NIPS}, 5998--6008.

\bibitem[{Veyseh(2016)}]{veyseh2016cross}
Veyseh, A. P.~B. 2016.
\newblock Cross-lingual Question Answering Using Common Semantic Space.
\newblock In \emph{TextGraphs}, 15--19.

\bibitem[{Veyseh et~al.(2020{\natexlab{a}})Veyseh, Dernoncourt, Nguyen, Chang,
  and Celi}]{veyseh2020acronym}
Veyseh, A. P.~B.; Dernoncourt, F.; Nguyen, T.~H.; Chang, W.; and Celi, L.~A.
  2020{\natexlab{a}}.
\newblock Acronym Identification and Disambiguation shared tasks for Scientific
  Document Understanding.
\newblock In \emph{AAAI Workshop on Scientific Document Understanding}.

\bibitem[{Veyseh et~al.(2020{\natexlab{b}})Veyseh, Dernoncourt, Tran, and
  Nguyen}]{dataset}
Veyseh, A. P.~B.; Dernoncourt, F.; Tran, Q.~H.; and Nguyen, T.~H.
  2020{\natexlab{b}}.
\newblock What Does This Acronym Mean? Introducing a New Dataset for Acronym
  Identification and Disambiguation.
\newblock In \emph{COLING}, 3285--3301.

\bibitem[{Wolf et~al.(2020)Wolf, Debut, Sanh, Chaumond, Delangue, Moi, Cistac,
  Rault, Louf, Funtowicz, Davison, Shleifer, von Platen, Ma, Jernite, Plu, Xu,
  Scao, Gugger, Drame, Lhoest, and Rush}]{wolf-etal-2020-transformers}
Wolf, T.; Debut, L.; Sanh, V.; Chaumond, J.; Delangue, C.; Moi, A.; Cistac, P.;
  Rault, T.; Louf, R.; Funtowicz, M.; Davison, J.; Shleifer, S.; von Platen,
  P.; Ma, C.; Jernite, Y.; Plu, J.; Xu, C.; Scao, T.~L.; Gugger, S.; Drame, M.;
  Lhoest, Q.; and Rush, A.~M. 2020.
\newblock Transformers: State-of-the-Art Natural Language Processing.
\newblock In \emph{EMNLP}, 38--45.

\bibitem[{Xu et~al.(2020)Xu, Qiu, Zhou, and Huang}]{xu2020improving}
Xu, Y.; Qiu, X.; Zhou, L.; and Huang, X. 2020.
\newblock Improving BERT Fine-Tuning via Self-Ensemble and Self-Distillation.
\newblock \emph{arXiv preprint arXiv:2002.10345} .

\bibitem[{Zhang et~al.(2019)Zhang, Zhang, Lu, Zhu, and Dong}]{yopo}
Zhang, D.; Zhang, T.; Lu, Y.; Zhu, Z.; and Dong, B. 2019.
\newblock You Only Propagate Once: Accelerating Adversarial Training via
  Maximal Principle.
\newblock In \emph{NeurIPS}, 227--238.

\bibitem[{Zhong et~al.(2021)Zhong, Zeng, Zhu, Zhang, Lin, Chen, and
  Tang}]{qwzhong2020}
Zhong, Q.; Zeng, G.; Zhu, D.; Zhang, Y.; Lin, W.; Chen, B.; and Tang, J. 2021.
\newblock Leveraging Domain Agnostic and Specific Knowledge for Acronym
  Disambiguation.
\newblock In \emph{AAAI Workshop on Scientific Document Understanding}.

\bibitem[{Zhu et~al.(2020)Zhu, Cheng, Gan, Sun, Goldstein, and Liu}]{free}
Zhu, C.; Cheng, Y.; Gan, Z.; Sun, S.; Goldstein, T.; and Liu, J. 2020.
\newblock FreeLB: Enhanced Adversarial Training for Natural Language
  Understanding.
\newblock In \emph{ICLR}.

\end{thebibliography}
